\pdfoutput=1

\documentclass[runningheads]{llncs}
\usepackage{graphicx}
\usepackage{amsmath,amssymb} 
\usepackage{color}\usepackage[width=122mm,left=12mm,paperwidth=146mm,height=193mm,top=12mm,paperheight=217mm]{geometry}

\usepackage{breakcites}

\usepackage{caption}
\usepackage[ruled,vlined]{algorithm2e}
\usepackage{float}
\usepackage{bm}
\usepackage{changepage}
\usepackage{subfigure}
\usepackage{multirow}

\begin{document}
\pagestyle{headings}
\mainmatter
\title{Weakly Supervised Learning of Objects, Attributes and their Associations} 

\titlerunning{Weakly Supervised Learning of Objects, Attributes and their Associations}

\authorrunning{Zhiyuan Shi, Yongxin Yang, Timothy M. Hospedales, Tao Xiang}

\author{Zhiyuan Shi, Yongxin Yang, Timothy M. Hospedales, Tao Xiang}
\institute{Queen Mary, University of London, London E1 4NS, UK {\tt\small \{z.shi,yongxin.yang,t.hospedales,t.xiang\}@qmul.ac.uk}}

\maketitle

\begin{abstract}

When humans describe images they tend to use combinations of nouns and adjectives, corresponding to objects and their associated attributes respectively. To generate such a description automatically, one needs to model objects, attributes and their associations. Conventional methods require strong annotation of object and attribute locations, making them less scalable. In this paper, we model object-attribute associations from weakly labelled images, such as those widely available on media sharing sites (e.g. Flickr), where only image-level labels (either object or attributes) are given, without their locations and associations. This is achieved by introducing a novel weakly supervised non-parametric Bayesian model. Once learned, given a new image, our model can describe the image, including objects, attributes and their associations, as well as their locations and segmentation. Extensive experiments on benchmark datasets demonstrate that our weakly supervised model performs at par with strongly supervised models on tasks such as image description and retrieval based on object-attribute associations.  
\end{abstract}

\section{Introduction}

Vision research is moving beyond simple classification, annotation and detection to encompass generating more structured and semantic descriptions of images. When humans describe images they use combinations of nouns and adjectives, corresponding to objects and their associated attributes respectively. For example, an image can be described as containing ``a person in red clothes and a shiny car''. In order to imitate this ability, a computer vision system needs to learn models about objects, attributes, and their associations. Object-attribute associations is important for avoiding the mistakes such as ``a shiny person and a red car''. Learning object-attribute association also provides new query capabilities, e.g., ``find images with a furry brown horse and a red shiny car''.

\begin{figure}
\centering
\includegraphics[width=\linewidth]{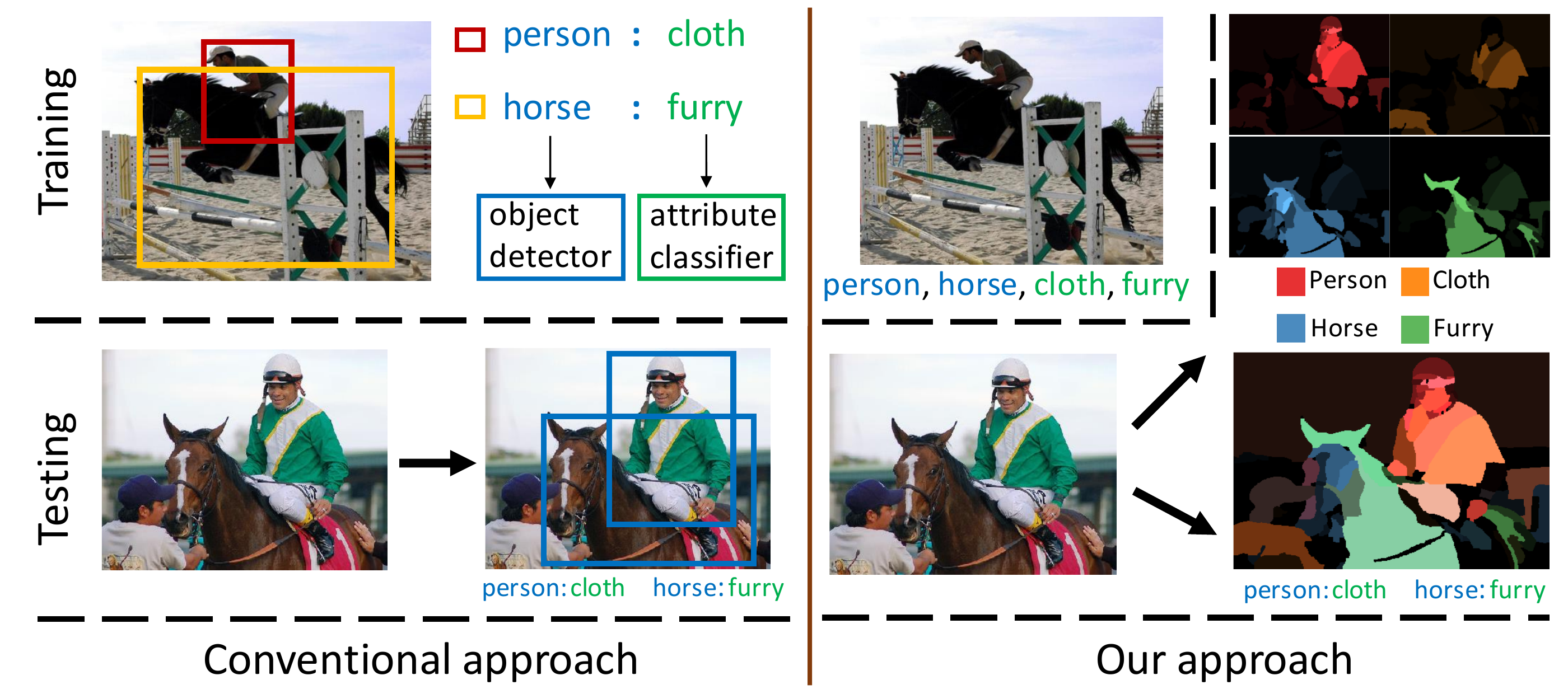}
\vskip -0.2cm
\caption{Comparing our weakly supervised approach to  object-attribute association learning to the conventional strongly supervised approach.}
\label{fig:schematic}
\end{figure}

There has been extensive work on detecting and segmenting objects \cite{socher2010sslModalities,deselaers2012wslGeneric,Shi_2013_ICCV} and describing specified objects and images in terms of semantic attributes \cite{Farhadi09CVPR,wang2013wslAttrLoc,bourdev2011attribposelet,siddiquie2011img_attrib_query}. 
However, these tasks have previously been treated separately;  jointly learning about and inferring object-attribute association in images with potentially multiple objects is much less studied.
The few existing studies on modelling object-attribute association \cite{kulkarni2011descriptions,Wang_2010_ECCV,Mahajan_2011_ICCV,Wang_2013_ICCV,wang2013wslAttrLoc,wang2009attrib_class_sal} use fully annotated data \cite{Wang_2010_ECCV,Mahajan_2011_ICCV,Wang_2013_ICCV} and/or are separately trained \cite{kulkarni2011descriptions,wang2009attrib_class_sal}. In the conventional pipeline (Fig.~\ref{fig:schematic}) images are strongly labelled with object bounding boxes and associated attributes, from which object detectors and attribute classifiers are trained.Given a new image,  the learned object detectors are first applied to find object locations, where the attribute classifiers are then applied to produce the object descriptions. However, there is a critical limitation of the existing approach: it requires strongly labelled objects and attributes. Considering there are over 30,000 object classes distinguishable to humans \cite{lampert13AwAPAMI}, even more attributes to describe them, and an infinite number of combinations, it is not scalable.   

In this paper  we propose to learn objects, attributes, and their associations from weakly labelled data. That is, images with object and attribute labels but not their associations nor their locations (see Fig.~\ref{fig:schematic}). Such weakly labelled images are abundant on  media sharing websites such as Flickr. Therefore lack of training data would never be a problem. However, learning strong semantics, i.e.~explicit object-attribute association from weakly labelled images is extremely challenging due to the label ambiguity: a real-world image with the tags ``dog, white, coat, furry'' could contain a furry dog and a white coat or a furry coat and a white dog. Furthermore, the tags/labels typically only describe the foreground/objects. There could be a white building in the background which is ignored by the annotator, and a computer vision model must infer that this is not what the tag `white' refers to. Conventional methods cannot be applied without object locations and explicit object-attribute association being labelled. 

To address the challenges of learning strong semantics from weak annotation, we develop a unified probabilistic generative model capable of jointly learning objects, attributes and their associations, as well as their location and segmentation. Our model is also able to learn from realistic images where there are multiple objects of variable sizes per image such as PASCAL VOC. More specifically, our model generalises the non-parametric Indian Buffet Process (IBP) \cite{Griffiths_2011_JMLR}. The IBP is chosen because it is designed for explaining multiple factors that simultaneously co-exist to account for the appearance of a particular image or patch, e.g., such factors can be an object and its particular texture and colour attributes. However, the conventional IBP is limited in that it is unsupervised and, as a flat model, applies to either patches or images, not both; it thus cannot be directly applied to our problem. To overcome these limitations, a novel model termed Weakly Supervised Stacked Indian Buffet Process (WS-SIBP) is formulated in this work. By introducing hierarchy into IBP, WS-SIBP is able to group data, thus allowing it to explain images as groups of patches, each of which has an inferred multi-label description vector corresponding to an object and its associated attributes. We also introduce weak image-level supervision, which is disambiguated into multi-label patch explanations by our WS-SIBP.

Modelling weakly labelled images using our framework provides a number of benefits: (i) By jointly learning multiple objects, attributes and background clutter in a single framework, ambiguity in each is explained away by knowledge of the other. (ii) The infinite number of factors provided by the non-parametric Bayesian framework allows structured background clutter of unbounded complexity to be explained away. (iii) A sparse binary latent representation of each patch allows an unlimited number of attributes to co-exist on one object.  The aims and capabilities of our approach are illustrated schematically in Fig.~\ref{fig:schematic}, where weak annotation in the form of a mixture of objects and attributes is transformed into object and attribute associations with locations.   

\section{Related work}
\label{sec:related work}

\noindent \textbf{Learning objects and attributes}\quad \noindent A central task in computer vision is  understanding image content. Such an understanding has been shown in the form of an image description in terms of nouns (object detection or region segmentation), and more recently adjectives (visual attributes) \cite{Farhadi09CVPR,RussakovskyECCV10}. Attributes have been used to describe objects \cite{Farhadi09CVPR,Turakhia_2013_ICCV}, people \cite{bourdev2011attribposelet}, clothing \cite{Chen_2013_ECCV}, scenes \cite{wang2013wslAttrLoc}, faces \cite{siddiquie2011img_attrib_query}, and video events \cite{fu2012attribsocial}.  However, most previous studies have  learned and inferred object and attribute models separately, e.g., by independently training  binary classifiers, and require strong annotations/labels indicating object/attribute locations and/or associations if the image is not dominated by a single object. 

\noindent \textbf{Learning object-attribute associations}\quad   
A few recent studies have learned object-attribute association explicitly \cite{wang2009attrib_class_sal,Kovashka_2011_ICCV,kulkarni2011descriptions,wang2013wslAttrLoc,Wang_2013_ICCV,Wang_2010_ECCV,Mahajan_2011_ICCV}. Different from our approach, \cite{wang2009attrib_class_sal,Wang_2013_ICCV,Wang_2010_ECCV,Mahajan_2011_ICCV} only trains and tests on unambiguous data, i.e. images containing a single dominant object, assumes object-attribute association is known at training; and moreover allocates exactly one attribute per object.  \cite{kulkarni2011descriptions} tests on more challenging PASCAL VOC data with multiple objects and attributes coexisting. However, their model is pre-trained on object and attribute detectors learned on strongly annotated images with object bounding boxes provided. \cite{wang2013wslAttrLoc} also does object segmentation and object-attribute prediction. But their model is learned from strongly labelled images in that object-attribute association are given during training; and importantly prediction is restricted to object-attribute pairs seen during training. In summary none of the existing work learns object-attribute association from weakly labelled data as we do here.

\noindent \textbf{Multi-attribute query}\quad Some existing work aims to perform attribute-based query \cite{Rastegari_CVPR13,Kovashka_2013_ICCV,multiattrs_cvpr2012,siddiquie2011img_attrib_query}. In particular, Recent studies  have considered how to calibrate \cite{multiattrs_cvpr2012} and fuse \cite{Rastegari_CVPR13} multiple attribute scores in a single query. We go beyond these studies in supporting conjunction of object+multi-attribute query. Moreover, existing methods either require bounding boxes or assume simple data with single dominant objects, and do not reason jointly about multiple attribute-object association. This means they would be intrinsically challenged in reasoning about (multi)-attribute-object queries on challenging data with multiple objects and multiple attributes in each image (e.g., querying furry brown horse, in a dataset with black horses and furry dogs in the same image). In other words, they cannot be directly extended to solve query by object-attribute association.

\noindent \textbf{Probabilistic models for image understanding}\quad Discriminative kernel methods underpin many high performance  recognition and annotation studies \cite{Farhadi09CVPR,RussakovskyECCV10,bourdev2011attribposelet,wang2013wslAttrLoc,sadeghi2011recoPhrases,kulkarni2011descriptions,chen2013neil,marchesotti2013beautUglyAttrs,deselaers2012wslGeneric}. However the flexibility of generative probabilistic models has seen them successfully applied to a variety of tasks, especially learning structured scene representations, and weakly-supervised learning \cite{fu2012attribsocial,socher2010sslModalities,Shi_2013_ICCV,LiSocherFeiFei2009}.  These studies often generalise  probabilistic topic models (PTM) \cite{blei2003lda}. However PTMs are limited for explaining objects and attributes in that latent topics are competitive - the fundamental assumption is that an object is a horse \emph{or} brown \emph{or} furry. They intrinsically do not account for the reality that it is \emph{all} at once.

We therefore generalise instead the Indian Buffet Process (IBP) \cite{velez2009variational,Griffiths_2011_JMLR}. The IBP is a latent feature model  that can independently activate each latent factor, explaining imagery as a weighted sum of active factor appearances. However, conventional IBP is (i) fully unsupervised, and (ii) only handles flat data. Thus, it could explain patches or images, but not images composed of patches, thereby limiting usefulness for multiple object-attribute association within images. We therefore formulate a novel  Weakly Supervised Stacked Indian Buffet Process (WS-SIBP) to model grouped data (images composed of patches), such that each patch has an infinite latent feature vector. This allows us to exploit image-level weak supervision, but disambiguate it to determine the best explanation in terms of which patches correspond to un-annotated background; which patch corresponds to which annotated object; and which objects have which attributes.

\noindent \textbf{Weakly supervised learning}\quad Weakly supervised learning (WSL) has attracted increasing attention as the volume of data which we are interested in learning from grows much faster than available annotation. Existing studies have generally focused on WSL of objects alone \cite{Shi_2013_ICCV,LiSocherFeiFei2009,deselaers2012wslGeneric}, with limited work on WSL of attributes \cite{wang2013wslAttrLoc,fu2012attribsocial}. Some studies have treated this as a discriminative multi-instance learning (MIL)  problem \cite{nguyen2010svm_miml,deselaers2012wslGeneric}, while others leveraged PTMs \cite{Shi_2013_ICCV,LiSocherFeiFei2009}. Weakly supervised localisation is a particularly challenging variant where images are annotated with objects, but absent bounding boxes means their location is unknown. This has been solved by sampling bounding boxes for MIL treatment \cite{deselaers2012wslGeneric}, or more `softly' by PTMs \cite{Shi_2013_ICCV}. In this paper we uniquely consider WSL of both objects, attributes, their associations and their locations simultaneously.

\noindent \textbf{Our contributions} \quad In this paper we make three key contributions: (i) We for the first time jointly learn all object, attribute and background appearances, object-attribute association, and their locations from realistic weakly labelled images; (ii) We formulate a novel  weakly supervised non-parametric Bayesian model by generalising the Indian Buffet Process; (iii) From this weakly labelled data, we demonstrate various image description and query tasks, including challenging tasks relying on predicting strong object-attribute association. Extensive experiments on benchmark datasets demonstrate that in each case our model is comparable to the strongly supervised alternatives and significantly outperforms a number of weakly supervised baselines.

\section{Weakly Supervised Stacked Indian Buffet Process}

We propose a non-parametric Bayesian model that learns to describe images composed of super-pixels/patches from weak object and attribute annotation. Each patch is associated with an infinite latent factor vector indicating if it corresponds to (an unlimited variety of) unannotated background clutter, or an object of interest, and what set of  attributes are possessed by the object.  Given a set of images with weak labels and segmented into super-pixels/patches, we need to learn: (i) which are the unique patches shared by all images with a particular label, (ii) which patches correspond to unannotated background, and (iii) what is the appearance of each object, attribute and background type. Moreover, since multiple labels (attribute and object) can apply to a single patch, we need to disambiguate which aspects of the appearance of the patch are due to each of the (unknown) associated object and attribute labels. To address all these learning tasks we build on the IBP \cite{velez2009variational} and  introduce a weakly-supervised stacked Indian Buffet process (WS-SIBP) to model data represented as bags (images) of instances (patches) with bag-level labels (image annotations). This is analogous to the notion of documents in topic models \cite{blei2003lda}.

\begin{figure}
\centering
\includegraphics[width=0.4\linewidth]{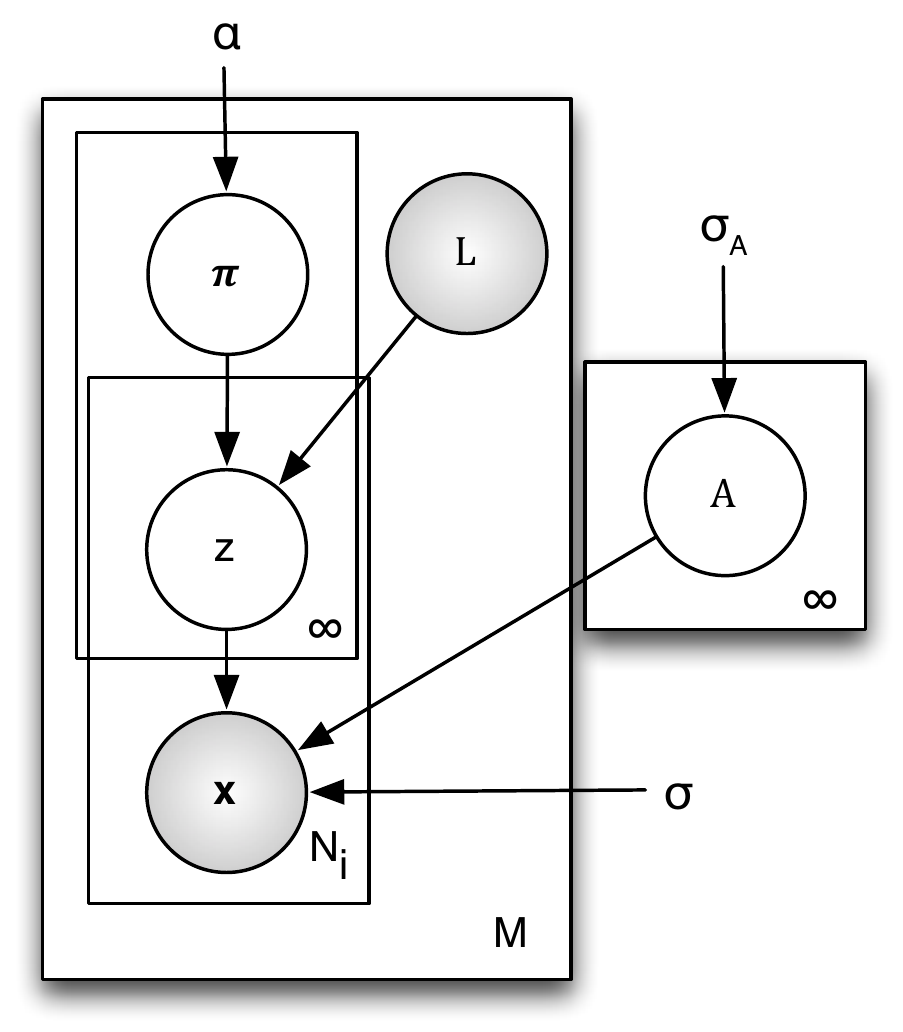}
\caption{The graphical model for WS-SIBP. Shaded nodes are observed. }
\label{fig:PGM}
\end{figure}

\subsection{Model formulation}

First, we associate each object category and each attribute to a latent factor. If there are $K_o$ object categories and $K_a$ attributes, then the first $K_{oa}=K_o+K_a$ latent factors correspond to these. An unbounded number of further factors are available to explain away background clutter in the data. At training time, we assume a 
 binary label vector $L^{(i)}$ for objects and attributes is provided for each image $i$. So $L^{(i)}_k=1$ if attribute/object $k$ is present, and zero otherwise. Also $L^{(i)}_k =1$ for all $k>K_{oa}$. That is, without any labels, we assume all background types can be present. With these assumptions, the generative process (illustrated in Fig.~\ref{fig:PGM}) for image $i$ represented as bags of patches $\mathbf{X}^{(i)}=\{\mathbf{X}^{(i)}_{j\cdot}\}$ is as follows:
\\~\\
\noindent For each latent factor $k\in1\dots \infty$:

\begin{enumerate}

\item Draw an appearance distribution mean $\mathbf{A}_{k\cdot} \sim \mathcal{N}(0,\sigma_{A}^2\bm{I})$.

\end{enumerate}
For each image $i\in1\dots M$:
\begin{enumerate}
\item Draw a sequence of i.i.d. random variables $v_1^{(i)},v_2^{(i)} \dots \sim \mbox{Beta}(\alpha, 1)$,

\item Construct an image prior $\pi_k^{(i)} = \prod\limits\limits_{t=1}^{k}v_t^{(i)}$,

\item Input weak annotation $L_k^{(i)} \in \{0, 1\}$,

\item For each super-pixel patch $j\in1\dots N_i$:

\begin{enumerate}

\item Sample state of each latent factor $k$: $z_{jk}^{(i)} \sim \mbox{Bern}(\pi_k^{(i)}L_k^{(i)})$,
\item Sample patch appearance: $\bm{X}_{j\cdot}^{(i)} \sim \mathcal{N}(\bm{Z}_{j\cdot}^{(i)}\bm{A},\sigma^2\bm{I}) $.

\end{enumerate}

\end{enumerate}

\noindent where $\mathcal{N}$, $\mbox{Bern}$ and $\mbox{Beta}$ respectively correspond to Normal, Bernoulli and Beta distributions with the specified parameters; and the notation $\bm{X}_{j\cdot}$ means the vector of row $j$ in matrix $\bm{X}$. The Beta-Bernoulli and Normal-Normal conjugacy are chosen because they allow more efficient inference. $\alpha$ is the prior expected sparsity of annotations and $\sigma^2$ is the prior variance in appearance for each factor. 

Denote hidden variables by $\bm{H} = \{\bm{\pi}^{(1)},\dots,\bm{\pi}^{(M)}, \bm{Z}^{(1)},\dots, \bm{Z}^{(M)}, \bm{A}\}$, images by $\bm{X} = \{\bm{X}^{(1)},\dots,\bm{X}^{(M)}\}$, and parameters by $\bm{\Theta}=\{\alpha, \sigma_A, \sigma, \bm{{L}}\}$. Then the joint  probability of the variables and data given the parameters is:

\begin{eqnarray}
 p(\bm{H}, \bm{X}|\bm{\Theta}) & = & \prod_{i=1}^{M} \bigg(\prod_{k=1}^{\infty}\Big(p(\pi_k^{(i)}|\alpha)\prod_{j=1}^{N_i}p(z_{jk}^{(i)}|\pi_k^{(i)}, L_k^{(i)})\Big)\nonumber \\
 & &  \cdot\prod_{j=1}^{N_i}p(\bm{X}_{j\cdot}^{(i)}|\bm{Z}^{(i)}_{j\cdot}, \bm{A}, \sigma)\bigg)\prod_{k=1}^{\infty}p(\mathbf{A}_{k\cdot}|\sigma_{A}^2)\label{eq:joint}.
\end{eqnarray}

Learning in our model aims to compute the posterior $ p(\bm{H} | \bm{X}, \bm{\Theta})$ for: disambiguating and localising all the annotated ($L^{(i)}$) objects and attributes among the patches (inferring $\bm{Z}^{(i)}_{j\cdot}$), inferring the attribute and background prior for each image (inferring $\bm{\pi}^{(i)}$), and learning the appearance of each factor (inferring $\mathbf{A}_{k\cdot}$).

\subsection{Model learning}

Exact inference for $ p(\bm{H} | \bm{X}, \bm{\Theta})$ in our stacked IBP is intractable, so an approximate inference algorithm in the spirit of \cite{velez2009variational} is developed. The mean field variational approximation to the desired posterior $p(\bm{H} | \bm{X}, \bm{\Theta})$ is:
\begin{equation}
q(\bm{H})=\prod_{i=1}^{M}\big(q_{\bm{\tau}}(\bm{v}^{(i)})q_{\bm{\nu}}(\bm{Z}^{(i)})\big)q_{\bm{\phi}}(\bm{A})\label{eq:meanFieldVar}
\end{equation}

\noindent where $q_{\bm{\tau}}(v_k^{(i)}) = \mbox{Beta}(v_k^{(i)}; \tau_{k1}^{(i)} \tau_{k2}^{(i)})$, $q_{\bm{\nu}}(z_{jk}^{(i)}) = \mbox{Bernoulli}(z_{jk}^{(i)}; \nu_{jk}^{(i)})$, $q_{\bm{\phi}}(\bm{A}_{k\cdot}) = \mathcal{N}(\bm{A}_{k\cdot}; \bm{\phi}_k, \bm{\Phi}_k)$ and the infinite stick-breaking process for latent factors is truncated at $K_{max}$, so $\pi_k=0$ for $k>K_{max}$. 
A variational message passing (VMP) strategy \cite{velez2009variational} can be used to minimise the KL divergence of Eq.~(\ref{eq:meanFieldVar}) to the true posterior. Updates are obtained by deriving integrals of the form $\ln q(\mathbf{h})=E_{\mathbf{H}\backslash\mathbf{h}}\left[\ln p(\mathbf{H},\mathbf{X})\right] + C$ for each group of hidden variables $\mathbf{h}$.
These result in the series of iterative updates given in Algorithm \ref{alg:inference}, where $\varphi(\cdot)$ is the digamma function; and $q^{(i)}_{ms}$ and $\mathbb{E}_{\bm{v}}[\log(1-\prod\limits_{t=1}^{k}v_{t}^{(i)})]$ are given in \cite{velez2009variational}.
In practice, the truncation approximation means that our WS-SIBP runs with a finite number of factors $K_{max}$ where truncation factor $K_{max}$ can be freely set so long as it is bigger than the number of factors needed by both annotations and background clutter ($K_{bg}$), i.e., $K_{max} \gg K_o+K_a+K_{bg}$. Despite the combinatorial nature of the object-attribute association and localisation problem, our model is of complexity $\mathcal{O}(MNDK_{max})$ for $M$ images with $N$ patches, $D$ feature dimension and $K_{max}$ truncation factor.

\begin{algorithm}[t]
\SetAlgoLined
\While{not converge}
{
\For {k = 1 \emph{\KwTo} $K_{max}$}{
$\bm{\phi}_k =(\frac{1}{\sigma^2}\sum\limits_{i=1}^{M}\sum\limits_{j=1}^{N_i}\nu_{jk}^{(i)}(\bm{X}_{j\cdot}^{(i)}-\sum\limits_{l:l \neq k} \nu_{jl}^{(i)}\bm{\phi}_l))(\frac{1}{\sigma_A^2}+\frac{1}{\sigma^2}\sum\limits_{i=1}^{M}\sum\limits_{j=1}^{N_i}\nu_{jk}^{(i)})^{-1}$ 

$\bm{\Phi}_k = \Big(\frac{1}{\sigma_A^2}+\frac{1}{\sigma^2}\sum\limits_{i=1}^{M}\sum\limits_{j=1}^{N_i}\nu_{jk}^{(i)}\Big)^{-1}\bm{I}$
}
\For{i = 1 \emph{\KwTo} $M$}
{
\For {k = 1 \emph{\KwTo} $K_{max}$}{
$\tau_{k1}^{(i)} = \alpha + \sum\limits_{m=k}^{K_{max}}\sum\limits_{j=1}^{N_i}\nu_{jm}^{(i)} + \sum\limits_{m=k+1}^{K_{max}}(N_i - \sum\limits_{j=1}^{N_i}\nu_{jm}^{(i)})(\sum\limits_{s=k+1}^{m}q_{ms}^{(i)})$

$\tau_{k2}^{(i)} = 1 + \sum\limits_{m=k}^{K_{max}}(N_i - \sum\limits_{j=1}^{N_i}\nu_{jm}^{(i)})q_{mk}^{(i)}$

\For {j = 1 \emph{\KwTo} $N_i$}{
$\eta = \sum\limits_{t=1}^{k}(\varphi(\tau_{t1}^{(i)}) -  \varphi(\tau_{t2}^{(i)})) - \mathbb{E}_{\bm{v}}[\log(1-\prod\limits_{t=1}^{k}v_{t}^{(i)})]$

$\quad \quad -\frac{1}{2\sigma^2}(\mbox{tr}(\bm{\Phi}_k)+\bm{\phi}_k\bm{\phi}_k^T-2 \bm{\phi}_k(\bm{X}_{j\cdot}^{(i)}-\sum\limits_{l:l \neq k} \nu_{jl}^{(i)}\bm{\phi}_l)^T)$

$\nu_{jk}^{(i)} = \frac{L_k^{(i)}}{1+e^{-\eta}}$ 

}
}
}
}
\caption{\label{alg:inference}Variational Inference for WS-SIBP}
\end{algorithm}

\subsection{Inference for test data}

At testing time, the appearance of each factor $k$, now modelled by sufficient statistics $\mathcal{N}(\bm{A}_{k\cdot}; \bm{\phi}_k, \bm{\Phi}_k)$, is assumed to be known (learned from the training data), while annotations for each test image $L^{(i)}_k$ will need to be inferred. Thus Algorithm~\ref{alg:inference} still applies, but without the appearance update terms and with $L^{(i)}_k=1~\forall k$, to reflect the fact that all the learned object, attribute, and background types could be present without any prior knowledge.

\subsection{Applications of the model}\label{sec:postproc}

Given the learned model applied to testing data, we can perform the following tasks: 
\textbf{Free Annotation: } This is to describe an image using a list of nouns and adjectives corresponding to objects and their associated attributes, as well as locating them. To infer what objects are present in  image $i$, the first $K_o$ latent factors of the inferred $\bm{\pi}^{(i)}$ are thresholded or ranked to obtain a list of objects. This is followed by locating them via searching for the patches $j^*$ maximising $\mathbf{Z}^{(i)}_{jk}$, then thresholding or ranking the $K_a$ attribute latent factors in $\mathbf{Z}^{(i)}_{j^*k}$ to describe them. 

\textbf{Annotation given object names:} This is a more constrained variant of the free annotation task above. Given a named (but not located) object $k$, its associated attributes can be estimated by first finding the location as $j^*=\underset{j}{\arg\max} \  \bm{Z}^{(i)}_{jk}$, then the associated attributes  by $\bm{Z}^{(i)}_{j^*k}$ for $K_o<k \leq K_o+K_a$. 
\textbf{Object+Attribute Query: } Images can be queried for a specified object-attribute conjunction $<k_o, k_a>$ by searching for $i^*,j^*=\underset{j}{\arg\max} \  \bm{Z}^{(i)}_{jk_o} \cdot \bm{Z}^{(i)}_{jk_a}$.

\section{Experiments}

\label{sec:exp}

\noindent \textbf{Datasets:} \\
Various object and attribute datasets are available such as aPascal, ImageNet, SUN \cite{Patterson_2012_CVPR} and AwA \cite{lampert13AwAPAMI}.  We use aPascal because it has multiple objects per image; and ImageNet due to sharing  attributes widely across categories.

\begin{figure}[t]

\begin{minipage}{0.6\textwidth}
\raggedright
  \includegraphics[width=1.0\linewidth]{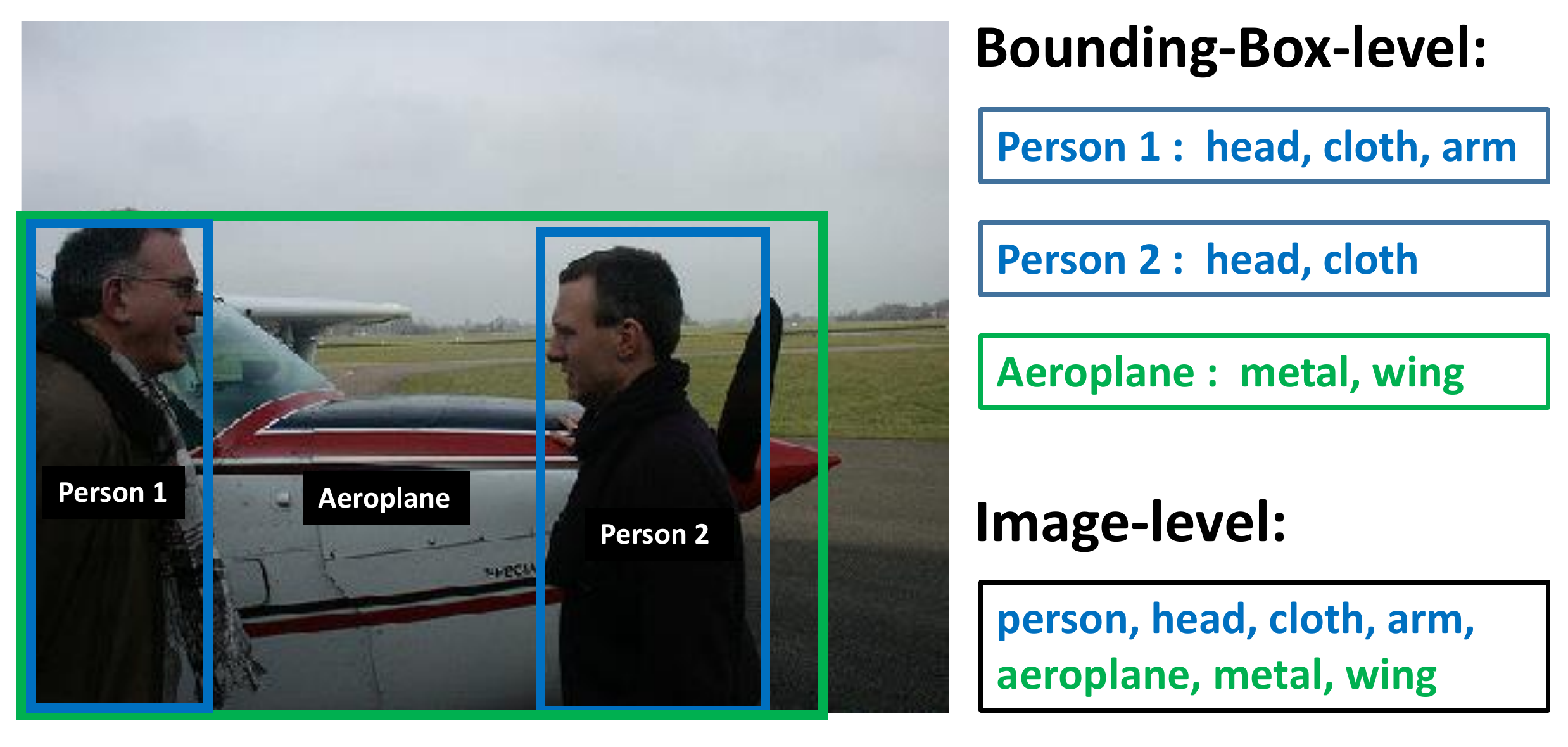}
  \captionof{figure}{Strong bounding-box-level annotation and weak image-level annotations for aPascal are used for learning strongly supervised models and weakly supervised models respectively.}
  \label{fig:dataset_annotat}
\end{minipage}%
\hspace{0.25cm}
\begin{minipage}{.34\textwidth}
\raggedleft
  \includegraphics[width=1.0\linewidth]{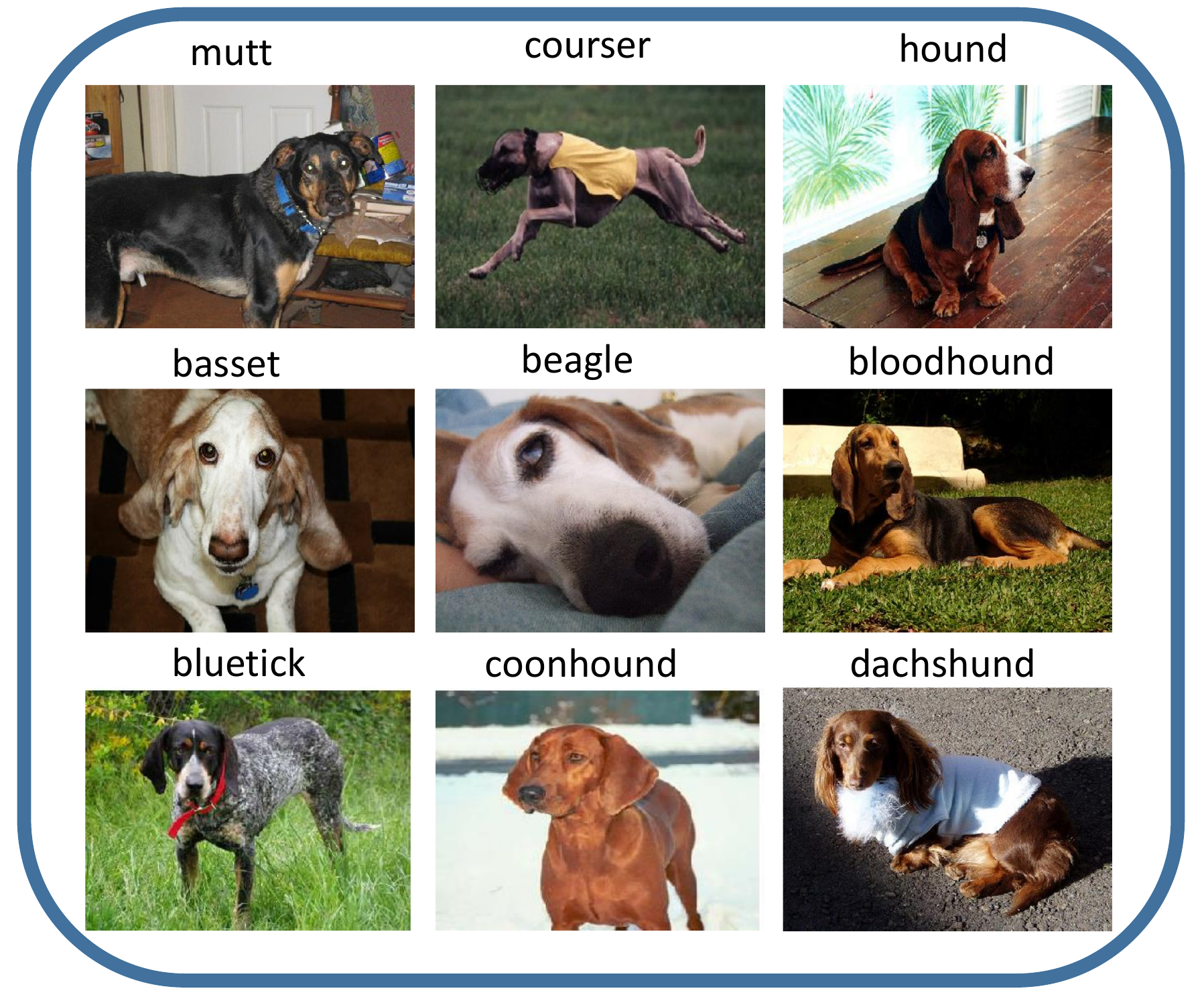}
  \captionof{figure}{43 subordinate classes of dog are converted into a single  entry-level class `dog'.}
   \label{fig:dataset_entry}
\end{minipage}
\end{figure}

\noindent\textbf{aPascal:}\quad This dataset \cite{Farhadi09CVPR} is an attribute labelled version of PASCAL VOC 2008. There are 4340 images of 20 object categories. Each object is annotated with a list of 64 attributes that describe them by shape (e.g., isBoxy), parts (e.g., hasHead) and material (e.g., isFurry). In the original aPascal, attributes are strongly labelled for 12695 object bounding boxes, i.e.~the object-attribute association are given. To test our weakly supervised approach, we merge the object-level category annotations and attribute annotations into a single annotation vector of length 84 for the entire image. This image-level annotation is much weaker than the original bounding-box-level annotation, as shown in Fig.~\ref{fig:dataset_annotat}. In all experiments, we use the same train/test splits provided by \cite{Farhadi09CVPR}. 

\noindent\textbf{ImageNet Attribute:} This dataset \cite{RussakovskyECCV10} contains 9600 images from 384 ImageNet synsets/categories. We ignore the provided bounding box annotation. Attributes for each bounding box are labelled as 1 (presence), -1 (absence) or 0 (ambiguous). We use the same 20 of 25 attributes as \cite{RussakovskyECCV10} and consider 1 and 0 as positive examples. Many of the 384 categories are subordinate categories, e.g.~dog breeds. However, distinguishing  fine-grained subordinate categories is beyond the scope of this study. We are interested in finding a `black-dog' or `white-car', rather than `black-mutt' or `white-ford-focus'. We thus convert the 384 ImageNet categories to 172 entry-level categories using \cite{Ordonez_2013} (see Fig.~\ref{fig:dataset_entry}). We evenly split each class to create the training and testing sets.

\noindent \textbf{Features:} \\
We first convert each image $i$ to $N_i$ super-pixels/patches by a recent segmentation algorithm \cite{amfm_pami2011}. We set the segmentation threshold to 0.1 to obtain a single over-segmentation from the hierarchical segmentation for each image. Each segmented patch is represented using two types of normalised histogram features: SIFT and Color. (1)  SIFT: we extract regular grid (every 5 pixels) colorSIFT \cite{vandeSandeTPAMI2010} at four scales. A 256 component GMM  model is constructed on the collection of ColourSIFTs from all images. We compute Fisher Vector + PCA for all regular points in each patch following \cite{Aggregating_TPAMI_2011}. The resulting reduced descriptor is 512-D for every segmented region. (2)  Colour: We convert the image to quantised LAB space 8$\times$8$\times$8. A 512-D color histogram is then computed for each patch. The final 1024-D feature vector concatenates SIFT and Colour features together.

\noindent \textbf{Compared Methods:}\\
 We compare our WS-IBP to one strongly supervised model and three weakly supervised alternatives:

\noindent\textbf{Strongly supervised model}: A strongly supervised model uses bounding-box-level annotation. Two variants are considered for the two datasets respectively. \textbf{DPM+s-SVM}: for aPascal, both object detector and attribute classifier are trained from fully supervised data (i.e.~Bounding-Box-level annotation in Fig.~\ref{fig:dataset_annotat}). Specifically, we use the  20 pre-trained DPM detectors from \cite{lsvm_pami} and 64 attribute classifiers from \cite{Farhadi09CVPR}. \textbf{GT+s-SVM}: for ImageNet attributes, there is not enough data to learn 172 strong  DPM detectors as in aPascal.  So we use the ground truth bounding box instead assuming we have perfect object detectors, giving a significant advantage to this strongly supervised model. We train attribute classifiers using our feature and liblinear SVM \cite{REF08a}. These strongly supervised models are similar in spirit to the models used in \cite{kulkarni2011descriptions,wang2013wslAttrLoc,wang2009attrib_class_sal} and can provide a performance upper bound for the weakly supervised models compared.

\noindent\textbf{w-SVM} \cite{Farhadi09CVPR,RussakovskyECCV10}:~~In this weakly-supervised baseline, both object detectors and attribute classifiers are trained on the weak image-level labels as for our model (see Fig.~\ref{fig:dataset_annotat}). For aPascal, we train object and attribute classifiers using the feature extraction and model training codes (which is also based on \cite{REF08a}) provided by the authors of \cite{Farhadi09CVPR}. For ImageNet, our features are used, without segmentation.

\noindent\textbf{MIML} \cite{Zhou20122291}:~~This is the multi-instance multi-label (MIML) learning method in \cite{Zhou20122291}. In a way, our model can also be considered as a MIML method with each image a bag and each patch an instance. The MIML model  provides a mechanism to use the same super-pixel/patch based representation for images as our model, thus providing the object/attribute localisation capability as our model does. 
 
\noindent\textbf{w-LDA}:~~Weakly-supervised Latent Dirichlet Allocation (LDA) approaches \cite{Rasiwasia_2013,Shi_2013_ICCV} have been used for object localisation. We implement a  generalisation of LDA \cite{blei2003lda,Shi_2013_ICCV} that accepts continuous feature vectors (instead of bag-of-words). Like MIML this method can also accept patch based representation, but w-LDA is more related to our WS-SIBP than MIML since it is also a generative model.

\subsection{Image annotation with object-attribute association}

An image description can be automatically generated by predicting objects and their associated attributes. Evaluating the performance of a multi-faceted framework covering annotation, association and localisation is non-trivial. To comprehensively cover all aspects of performance of our method and competitors, we perform three annotation tasks with different amount of constraints on test images: (1) \textit{free annotation}, where no constraint is given to a test image, (2) \textit{annotation given object names}, where named but not located objects are known for each test image, and  (3) \textit{annotation given locations}, where objects locations are given in the form of bounding boxes, where the attributes can be predicted.

\begin{table}[t]
\setlength{\tabcolsep}{0.2em}
\centering
{\footnotesize
\begin{tabular}{l | l  l  l |  l  | l  }

\hline
aPascal & w-SVM & MIML & w-LDA  & WS-SIBP  &  DPM+s-SVM  \\

\hline
AP@2    &  24.8 & 28.7 &  30.7 & 38.6 &  40.6 \\ 
AP@5   &   21.2 & 22.4 & 24.0 & 28.9 &  30.3 \\ 
AP@8  &    20.3  & 21.0 & 21.5 & 24.1 &  23.8 \\ 
\hline
ImageNet  & w-SVM  & MIML & w-LDA &  WS-SIBP  &  GT+s-SVM \\
\hline
AP@2 & 46.3 &   46.6 & 48.4& 58.5&65.9\\ 
AP@3  & 41.1 &   43.2  & 43.1& 51.8 &60.7 \\ 
AP@4 &  37.5 &   38.3 & 38.4& 47.4 & 53.2\\ 
\hline
\end{tabular}}
\caption{Free annotation performance evaluated on $t$ attributes per object.}
\label{tab:exp1_real}
\end{table}
\begin{figure}[t]
\centering
\includegraphics[width=1.0\linewidth]{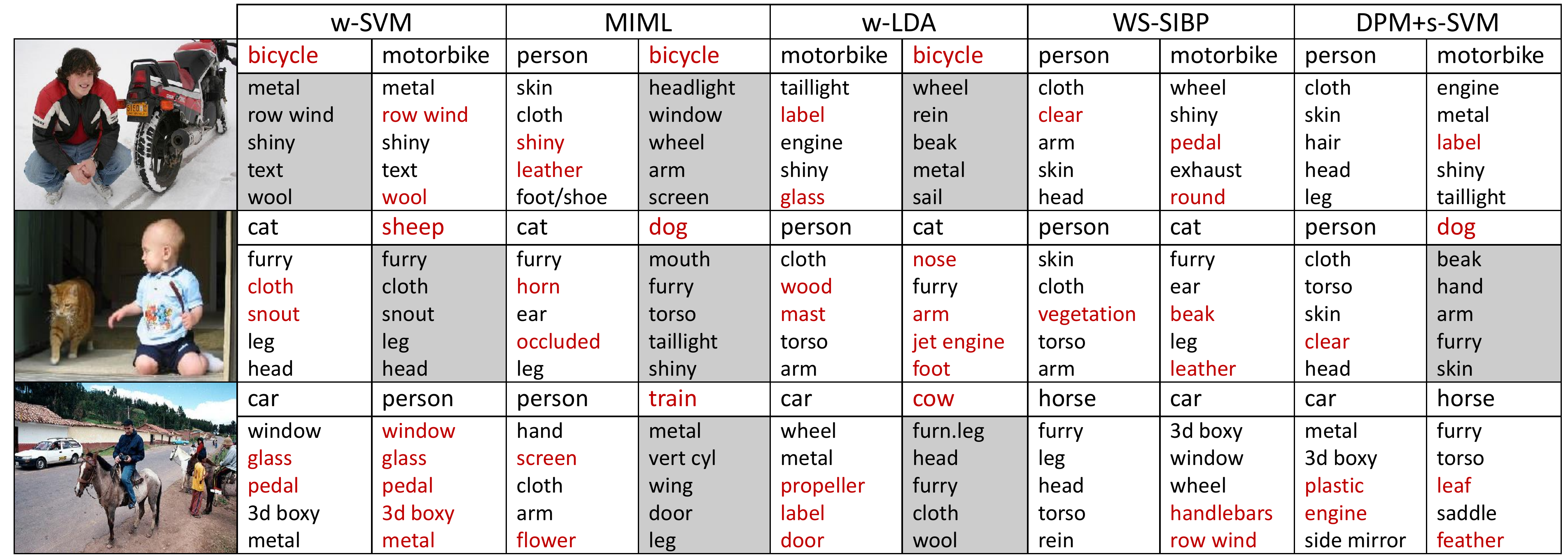}
\vskip -0.2cm
\caption{Qualitative results on free annotation. False positives are shown in red. If the object prediction is wrong, the corresponding attribute box is shaded. }
\label{fig:exp1_real_vis}
\end{figure}

\noindent \textbf{Free annotation:} For  WS-SIBP, w-LDA and MIML the procedure in Sec.~\ref{sec:postproc} is used to  detect objects and then describe them using the top $t$ attributes. 
For the strongly supervised model on aPascal (DPM+s-SVM), we use DPM object detectors to find the most confident objects and their bounding boxes in each test image. Then we use the 64 attribute classifiers to predict top $t$ attributes in each bounding box. In contrast, w-SVM trains attributes and objects independently, and cannot associate objects and attributes. We thus use it to predict only one attribute vector per image regardless of which object label it predicts. 

Since there are variable number of objects per image in aPascal, quantitatively evaluating free annotation is not straightforward. 
Therefore, we evaluate only the most confident object and its associated top $t$ attributes in each image, although more could be described. For ImageNet, there is only one object per image. 
We follow \cite{Feng_2013_ICCV,tag_2013} in evaluating annotation accuracy by average precision (AP), given varying numbers ($t$) of predicted attributes per object. Note that if the predicted object is wrong, all associated attributes are considered wrong.

Table \ref{tab:exp1_real} compares the free annotation performance of the five models. We have the following observations: (1) Our WS-SIBP, despite learned with the weak image-level annotation, yields comparable performance to the strongly supervised model. The gap is particularly small for the more challenging aPascal dataset, whist for ImageNet, the gap is bigger as the strongly supervised GT+s-SVM has an unfair advantage  by using the ground truth bounding boxes during testing. (2) WS-SIBP consistently outperforms the three weakly supervised alternatives. The margin is particularly large for $t=2$ attributes per object, which is closest to the true number of attributes per object. For bigger $t$, all models must generate some irrelevant attributes thus narrowing the gaps.  (3) As expected, the w-SVM model obtains the weakest results, suggesting that the ability to locate objects is important for modelling object-attribute association. (4) Compared to the two generative models, MIML has worse performance because a generative model is more capable of utilising weak labels \cite{Shi_2013_ICCV}. (5) Between the two generative models, the advantage of our WS-SIBP over w-LDA is clear; due to the ability of IBP to explain each patch with multiple non-competing factors. (Training two independent w-LDA models for objects and attributes respectively is not a solution: the problem would re-occur for multiple competing attributes.)

Fig.~\ref{fig:exp1_real_vis} shows qualitative results on aPascal via the two most confident objects and their associated attributes.
This is challenging data -- even the strongly supervised DPM+s-SVM makes mistakes for both attribute and object prediction. Compared to the weakly supervised models, WS-SIBP has more accurate prediction --  it jointly and non-competitively models objects and their attributes so object detection benefits from attribute detection and vice versa. 
Other weakly supervised  models are also more likely to mismatch attributes with objects, e.g.~MIML detects a shiny person rather than the correct shiny motorbike. 

\begin{figure}[t]
\centering
\includegraphics[width=1.0\linewidth]{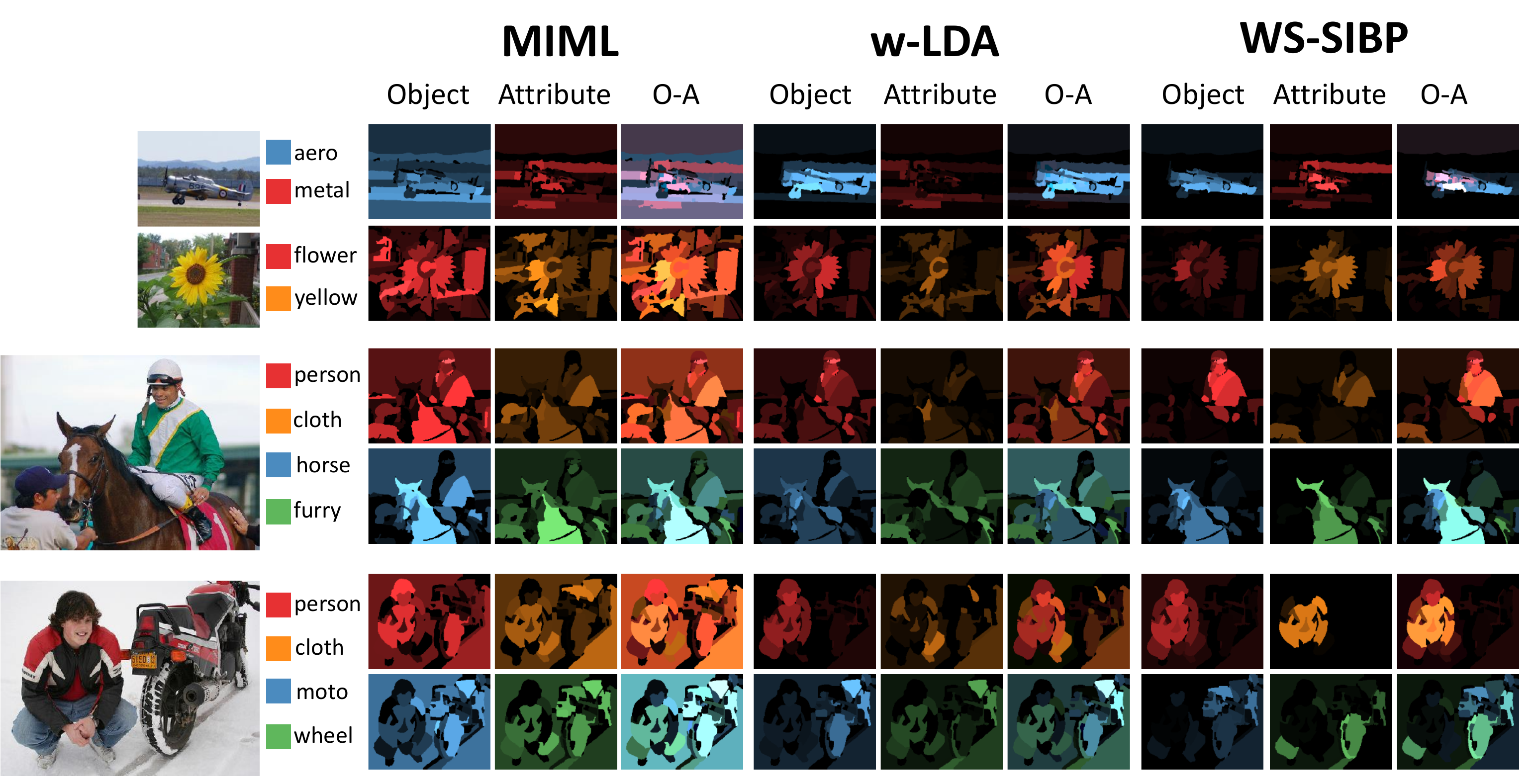}
\caption{Illustrating the inferred patch-annotation. Object and attributes are coloured, and multi-label annotation blends colours. The bottom two groups each have two rows corresponding to the two most confident objects detected.  }
\label{fig:obj_att_topic}
\end{figure}

To gain some insight into what has been learned by our model and why it is better than the weakly supervised alternatives, Fig.~\ref{fig:obj_att_topic}  visualises the attribute and object factors learned by  WS-SIBP model and by the two baselines that also use  patches as  input. It is evident that without explicit  background modelling, MIML suffers greatly by trying to explain the background patches using the weak labels. In contrast, both w-LDA and WS-SIBP have good segmentation of  foreground objects, showing that both the learned foreground and background topics are meaningful. However, for w-LDA, since object and attributes topics compete for the same patch, each patch is dominated by either an object  or attribute topic. In contrast, the object factors and attribute factors co-exist happily in WS-SIBP as they should do, e.g.~most person patches have the clothing attribute as well. 

\noindent \textbf{Annotation given object names (GN):} In this experiment, we assume that object labels are given and we aim to describe each object by attributes, corresponding to tasks such as: ``Describe the car in this image". 
For the strongly supervised model on aPascal, we use the object's DPM detector to find the most confident bounding box. Then we predict attributes for that box. Here,  annotation accuracy is the same as attribute  accuracy, so the performance of different models is evaluated following \cite{Zhang_2013_ICCV} by mean average precision (mAP) under the precision-recall curve. Note that for aPascal,  w-SVM reports the same list of attributes for all co-existing objects, without being able to localise and distinguish them. Its result is thus not meaningful and is excluded. The same set of conclusions can be drawn from Table~\ref{tab:att_pre_apascal} as in the free annotation task: our WS-SIBP at par with the supervised models and outperforming the  weakly supervised ones. 

\begin{table}[t]
\setlength{\tabcolsep}{0.2em}
\centering
\begin{tabular}{l|l | l  l  l | l | l  }
\hline
&&  w-SVM  & MIML & w-LDA  &  WS-SIBP   &  strongly supervised  \\
\hline \parbox[t]{3mm}{\multirow{2}{*}{\rotatebox[origin=c]{90}{GN}}} &aPascal   & -- & 32.1 & 35.5  & 38.9 & 41.8 \\ 
&ImageNet&  32.4  & 33.5  & 39.6 &   51.5 & 56.8\\ 
\hline
\parbox[t]{3mm}{\multirow{2}{*}{\rotatebox[origin=c]{90}{GL}}} &aPascal   &  33.2& 35.1 & 35.8  & 43.8  & 42.1   \\ 
&ImageNet & 37.7  &  39.1  &  46.8  &   53.7  &  56.8  \\ 
\hline
\end{tabular}
\caption{Results on annotation given object names (GN) or locations (GL). }

\label{tab:att_pre_apascal}
\end{table}
\begin{figure}[t]
\centering
\subfigure[O-A, aPascal]{
  \centering
  \includegraphics[height=.253\linewidth]{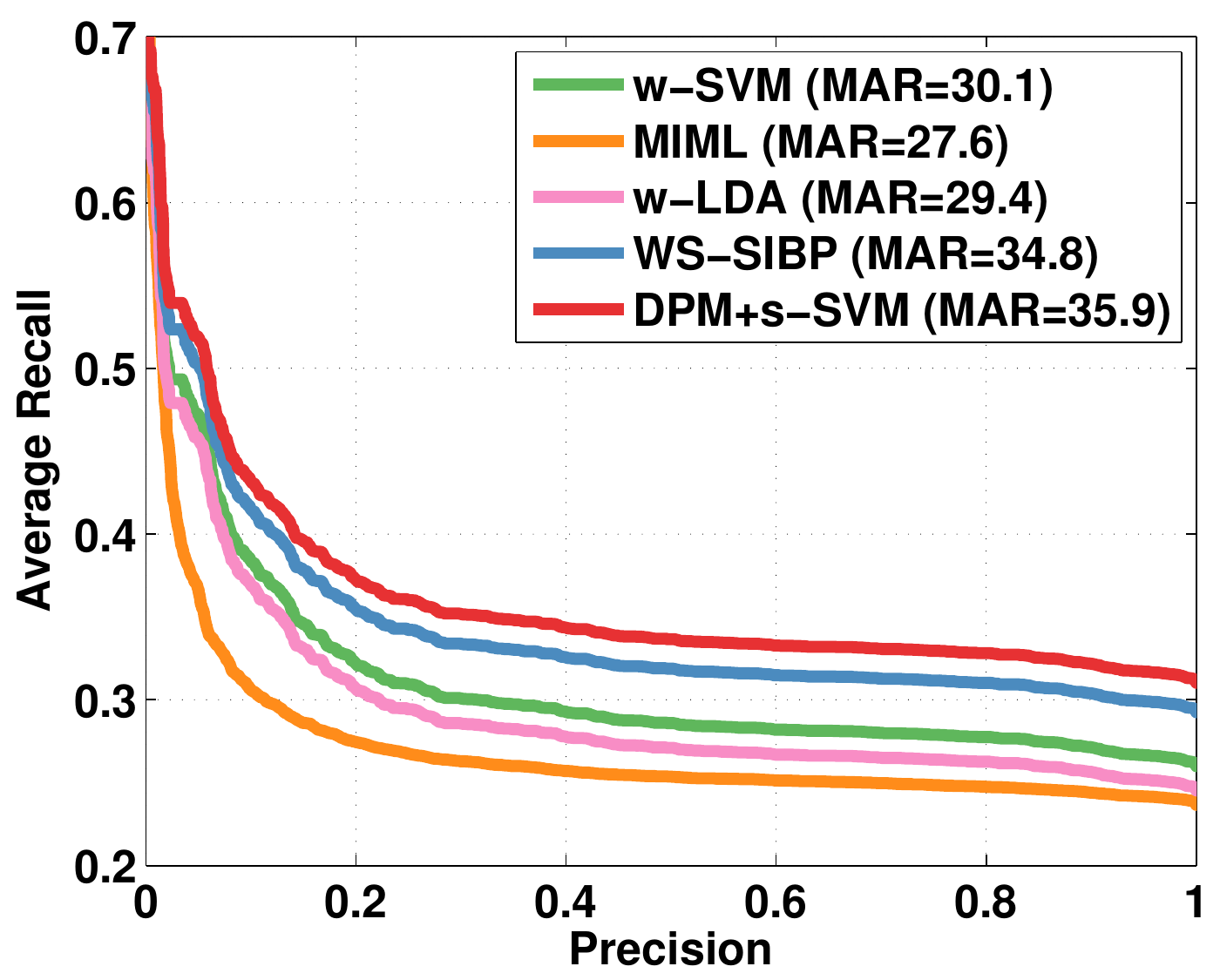}
  \label{fig:sub1}
}%
\subfigure[O-A, ImageNet]{
  \centering
  \includegraphics[height=.253\linewidth]{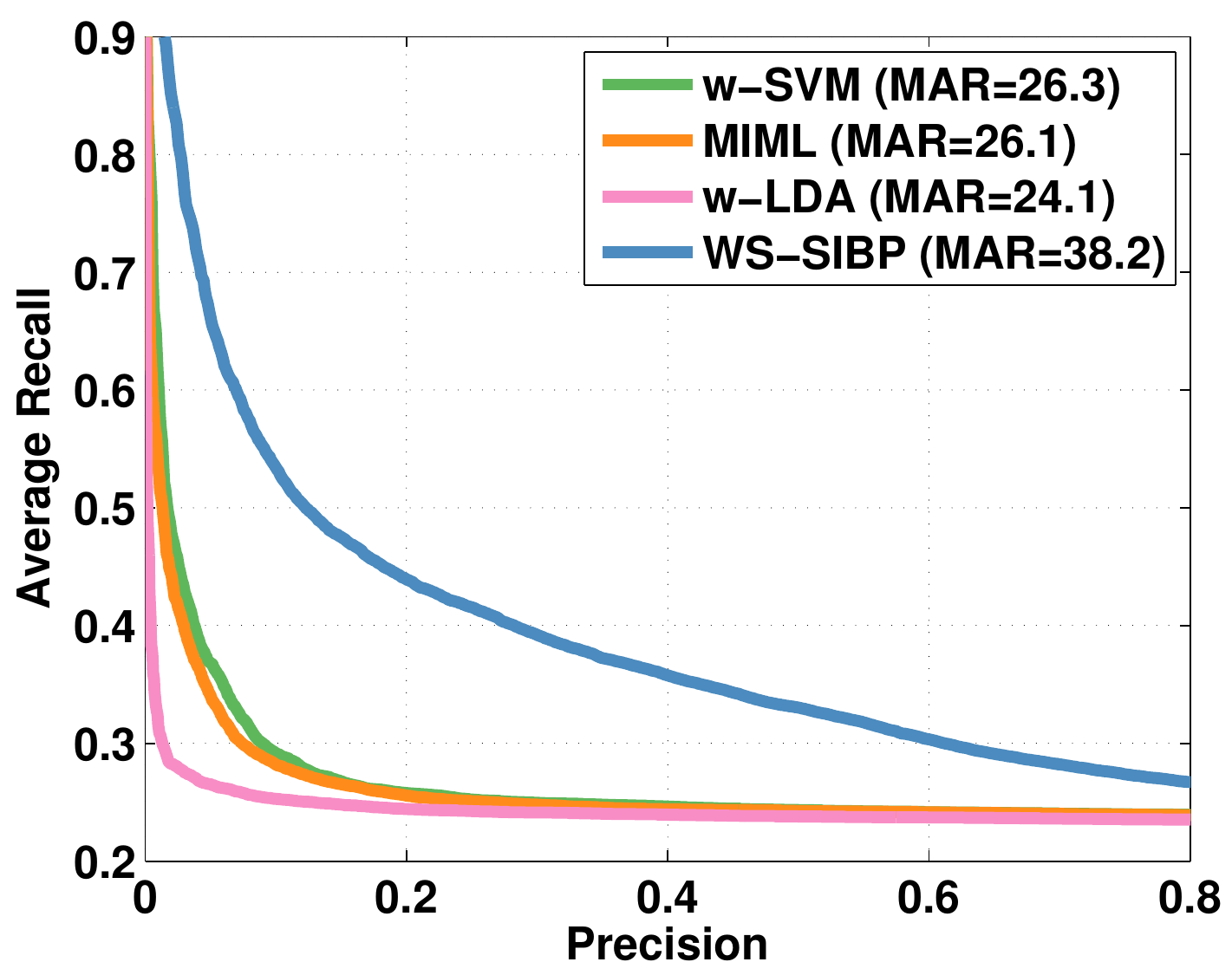}
  \label{fig:sub1}
}%
\subfigure[O-A-A, ImageNet]{
  \centering
  \includegraphics[height=.253\linewidth]{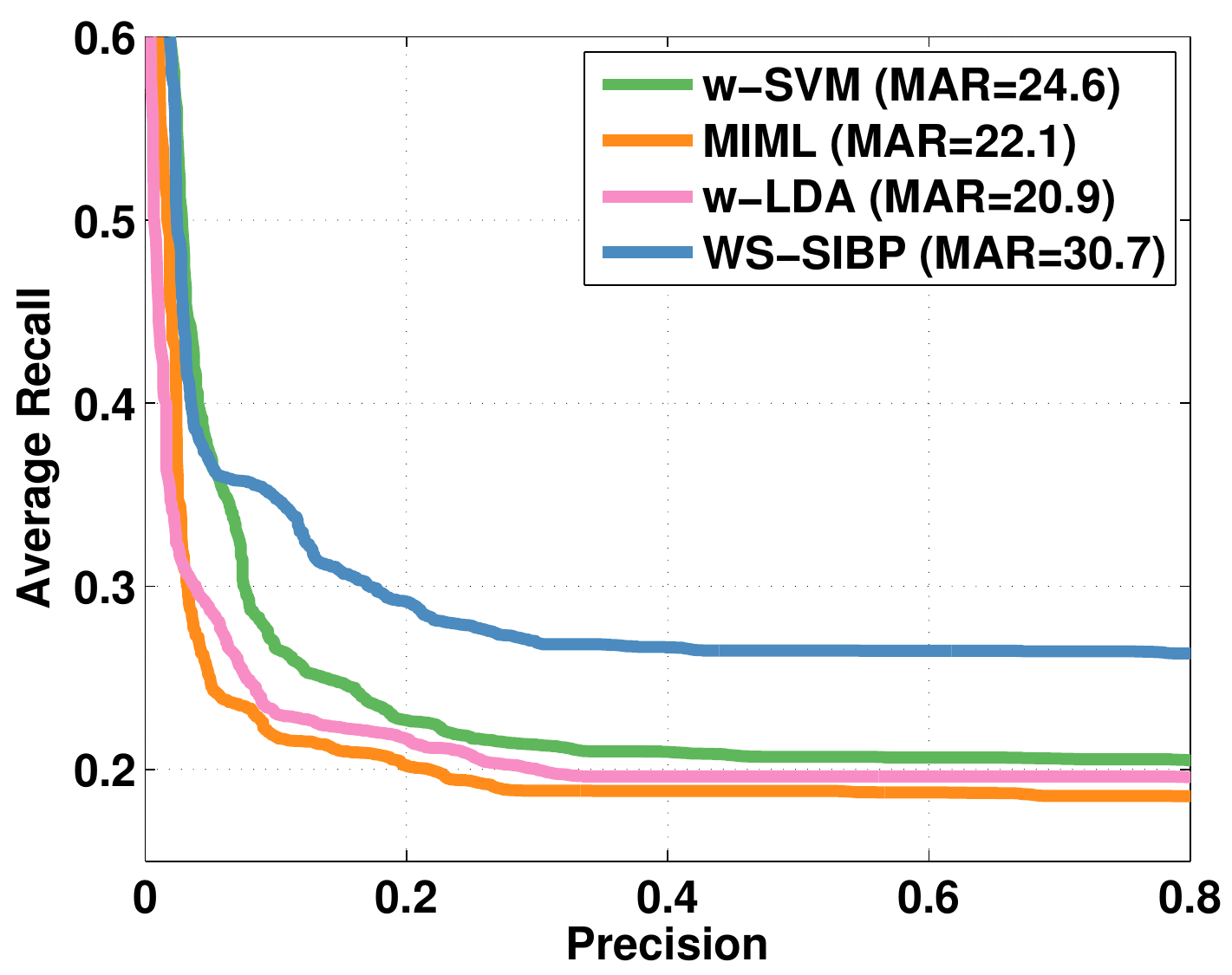}
  \label{fig:sub1}
}%
\caption{Object-attribute query results as precision-average recall curve.}
\label{fig:test}
\end{figure}

\noindent \textbf{Given object location (GL):} If we further know the bounding box of an object in a test image, we can simply predict attributes inside each bounding box. This becomes  the conventional attribute prediction task \cite{Farhadi09CVPR,RussakovskyECCV10} for describing an object. Table~\ref{tab:att_pre_apascal} shows the results, where similar observations can be made as in the other two tasks above. Note that in this case  the strongly supervised model  is the method used in \cite{Farhadi09CVPR}. The mAP obtained using our weakly supervised model is even higher than the strongly supervised model (though our area-under-ROC-curve value of 81.5 is slightly lower than the 83.4 figure reported in \cite{Farhadi09CVPR}).  

\subsection{Object-attribute query}

In this task object-attribute association is used for image retrieval.  Following work on multi-attribute queries \cite{Rastegari_CVPR13}, we use mean average recall over all precisions (MAR) as the evaluation metric. Note that unlike \cite{Rastegari_CVPR13} which requires each queried \emph{combination} to have enough (100) training examples to train conjunction classifiers, our method can query novel never-previously-seen combinations. Three experiments are conducted. We generate 300 random object-attribute combinations for aPascal and ImageNet respectively and 300 object-attribute-attribute queries for ImageNet. For the strongly supervised model, we normalise and multiply  object detector with attribute classifier scores. No object detector is trained for ImageNet so no result is reported there. For w-SVM, we use  \cite{multiattrs_cvpr2012} to calibrate the SVM scores for objects and attributes as in \cite{Rastegari_CVPR13}. For the three WS models, the procedure in Sec.~\ref{sec:postproc} is used to compute the retrieval ranking. 

\begin{figure}[t]
\centering
\includegraphics[width=0.95\linewidth]{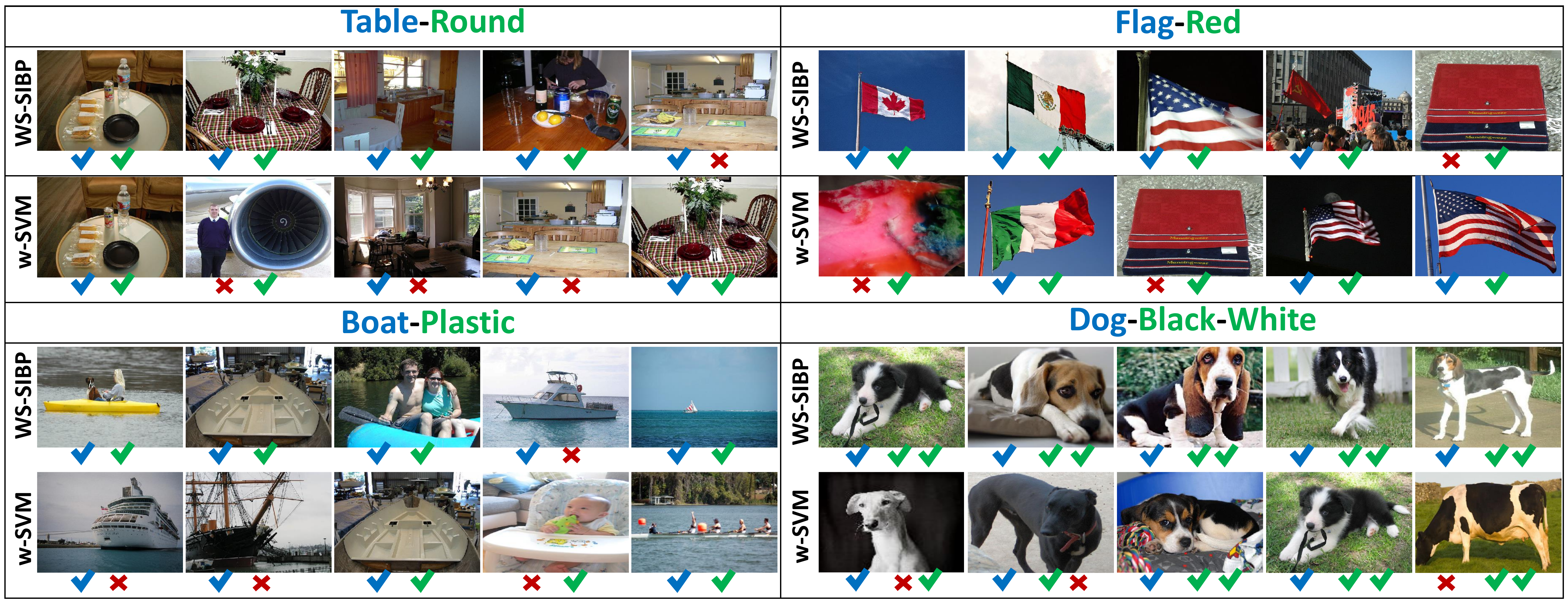}
\caption{Object-attribute query: qualitative comparison}
\label{fig:object_att_ann}
\end{figure}

Quantitative results are shown in Fig.~\ref{fig:test} and some qualitative examples  in Fig.~\ref{fig:object_att_ann}. Our WS-SIBP has a very similar MAR values to the strongly supervised DPM+s-SVM, while outperforming all the other models. 
w-SVM calibration  \cite{multiattrs_cvpr2012} helps it outperform MIML and w-LDA.
However, the lack of object-attribute association and background modelling still causes problems for w-SVM. This is illustrated in the `dog-black-white' example shown in Fig.~\ref{fig:object_att_ann} where a white background caused an image with a black dog retrieved at rank 2 by w-SVM.

\section{Conclusion}

We have presented an effective model for weakly-supervised learning of objects, attributes, their location and associations. Learning object-attribute association from weak supervision is non-trivial but critical for learning from `natural' data, and scaling to many classes and attributes. We achieve this for the first time through a novel weakly-supervised stacked IBP model that simultaneously disambiguates patch-annotation correspondence, as well as learning the appearance of each annotation. Our results show that our model performs comparably with a strongly supervised alternative that is significantly more costly to supervise.


\bibliographystyle{splncs03}
\bibliography{egbib}
\end{document}